\documentclass[letterpaper]{article} % DO NOT CHANGE THIS
\usepackage[preprint]{aaai2027}  % DO NOT CHANGE THIS
\usepackage[hyphens]{url}  % DO NOT CHANGE THIS
\usepackage{graphicx} % DO NOT CHANGE THIS
\usepackage{amsmath,amsfonts}
\usepackage{array}
\usepackage{textcomp}
\usepackage{verbatim}
\usepackage{graphicx}
\usepackage{cite}
\usepackage{booktabs}
\usepackage{xcolor}    
\usepackage{colortbl}
\usepackage{arydshln}
\usepackage{tabularx}

\usepackage{multirow}
\usepackage{makecell}
\usepackage{arydshln}
\usepackage{tikz}
\usepackage{multicol}
\usepackage{natbib}  % DO NOT CHANGE THIS AND DO NOT ADD ANY OPTIONS TO IT
\usepackage{caption} % DO NOT CHANGE THIS AND DO NOT ADD ANY OPTIONS TO IT
\usepackage{algorithm}
\usepackage{algorithmic}

\usepackage{newfloat}
\usepackage{listings}
\DeclareCaptionStyle{ruled}{labelfont=normalfont,labelsep=colon,strut=off} % DO NOT CHANGE THIS
\floatstyle{ruled}
\newfloat{listing}{tb}{lst}{}
\floatname{listing}{Listing}

\usepackage{booktabs}

\title{GaussFlaw: Fine-Grained Perceptual Diagnosis for 3D Gaussian Splatting}
\author{
    Shaocheng Shen,
    Zihan Zheng,
    Kejin Li,
    Lu Liu,
    Xiaoyun Zhang,\\
    Qiang Hu,
    Wenjun Zhang
}

\affiliations{
    Shanghai Jiao Tong University, Shanghai, China
}

\begin{document}

\maketitle

\begin{abstract}
3D Gaussian Splatting (3DGS) enables photorealistic novel-view synthesis with real-time rendering, yet fine-grained diagnosis of its rendering artifacts remains challenging. These artifacts are typically spatially localized and exhibit diverse degradation patterns, making them difficult to characterize using global quality metrics and image-level assessment benchmarks, which provide no explicit indication of \emph{where} they occur or \emph{what types} of artifacts they represent. To address this limitation, we introduce \textbf{GaussFlaw-80K}, the first large-scale dataset for fine-grained perceptual diagnosis in 3DGS. Constructed through a model-assisted human-in-the-loop pipeline, GaussFlaw-80K contains 80,049 artifact regions across 14,289 rendered images, each annotated with a pixel-level mask and one of eight flaw labels spanning geometric, blending, and photometric degradations. These annotations turn 3DGS artifact diagnosis into an explicit and measurable task. Building on this dataset, we propose \textbf{GeoCritic}, a geometry-guided multi-task model that integrates RGB appearance with depth-derived cues to jointly localize artifacts and classify their flaw types. Beyond evaluation, we further demonstrate the downstream benefits of fine-grained diagnosis through three complementary targeted operations: Gaussian densification, attribute correction, and outlier pruning. Experiments show that GeoCritic achieves an AUC-J of 0.9507 and an Acc@1 of 80.08\%, while diagnosis-guided refinement improves all reported global and crop-level metrics across diverse datasets and 3DGS backbones.
\end{abstract}

\section{Introduction}

\begin{table*}[t]
\centering
% 稍微调回1.0，保证两行表头不拥挤
\renewcommand\arraystretch{1.0} 
% 使用 resizebox 强行限制在 \textwidth 内
\resizebox{\textwidth}{!}{
\begin{tabular}{l c c c c c c c}
\toprule
% 保留两行表头设计，最大限度减小表格的自然宽度
Name & Scene & \begin{tabular}{@{}c@{}}Number of \\ Vids/Imgs\end{tabular} & \begin{tabular}{@{}c@{}}Number of \\ Ann.\end{tabular} & \begin{tabular}{@{}c@{}}Annotation \\ Format\end{tabular} & \begin{tabular}{@{}c@{}}Flaw \\ Location\end{tabular} & \begin{tabular}{@{}c@{}}Flaw \\ Type\end{tabular} & Refinement \\
\midrule
GSC-QA~\cite{10.1145/3696409.3700172} & 15 & 120 & 120 & Score & $\boldsymbol{\times}$ & $\boldsymbol{\times}$ & $\boldsymbol{\times}$ \\
GS-QA~\cite{11219925} & 8  & 64 & 64 & Score & $\boldsymbol{\times}$ & $\boldsymbol{\times}$ & $\boldsymbol{\times}$ \\
NVS-QA~\cite{zhang2025evaluatinghumanperceptionnovel} & 13 & 65 & 130 & Score & $\boldsymbol{\times}$ & $\boldsymbol{\times}$ & $\boldsymbol{\times}$ \\
3DGS-IEval-15K~\cite{10.1145/3746027.3758206} & 10 & 15.2K & 15.2K & Score & $\boldsymbol{\times}$ & $\boldsymbol{\times}$ & $\boldsymbol{\times}$ \\
\midrule
\textbf{GaussFlaw-80K (Ours)} & \textbf{26} & \textbf{14.2K} & \textbf{80K} & \textbf{Pixel-level Diagnosis} & \textbf{$\boldsymbol{\checkmark}$} & \textbf{$\boldsymbol{\checkmark}$ (8 Types)} & \textbf{$\boldsymbol{\checkmark}$ (Targeted)} \\
\bottomrule
\end{tabular}
}
\caption{\textbf{Comparison of 3DGS quality assessment datasets.} Compared to existing benchmarks, GaussFlaw-80K offers advantages in diversity and volume, featuring 26 complex scenes and 80K pixel-level annotations. By advancing the annotation format to fine-grained diagnosis (enabling explicit flaw location and typing), our dataset uniquely supports artifact analysis and targeted refinement.}
\label{tab:dataset_comparison}
\vspace{-10pt}
\end{table*}

% 3D Gaussian Splatting (3DGS)~\cite{kerbl20233d} has revolutionized the field of novel view synthesis with its exceptional real-time rendering speed and photorealistic quality. It has been widely adopted in core downstream tasks such as medical imaging \textbf{[cite]}, virtual reality \textbf{[cite]}, and digital twins \textbf{[cite]}. Despite achieving remarkable performance within the training distribution of dense views, existing 3DGS models persistently suffer from local artifacts, such as floaters, over-smoothing, and needle stretch. These localized flaws severely compromise human subjective perception, thereby hindering the transition of 3DGS into industrial applications. However, rectifying these artifacts requires pixel-level control and fine-grained geometric understanding, presenting a significant challenge for current implicit global optimization paradigms.
3D Gaussian Splatting (3DGS)~\cite{kerbl20233d} and its variants~\cite{feng2025flashgs,navaneet2023compact3d,hu20254dgc,NEURIPS2025_923285de} have emerged as powerful representations for novel-view synthesis, offering photorealistic rendering with real-time efficiency. Despite its impressive global quality, 3DGS renderings often remain locally unreliable: perceptually salient artifacts such as floaters, ghosting, needle-like stretching, and over-smoothed details can persist within otherwise realistic novel views. Importantly, these artifacts exhibit distinct spatial patterns and degradation characteristics, reflecting different failures in geometry, opacity and blending, or texture representation. Since such flaws are typically spatially localized, their perceptual impact can be easily obscured by the overall rendering quality. A reliable assessment of 3DGS therefore requires fine-grained perceptual diagnosis that explicitly answers two questions: \textbf{where} does an artifact occur, and \textbf{what} type of flaw does it represent?

% The fundamental bottleneck of this issue lies in the evaluation paradigm. Existing perceptual evaluations for 3DGS are predominantly confined to global full-reference metrics (e.g., PSNR, SSIM~\cite{1284395}, LPIPS~\cite{Zhang_2018_CVPR}), which fail to focus on localized flaws within otherwise high-quality renderings. Recent studies~\cite{11578166,11396862,10.1145/3696409.3700172} have explored multi-dimensional assessments: 3DGS-QA~\cite{6137fc1169544b478ea0253361365c08} focuses specifically on 3D static objects, 3DGS-IEval-15K~\cite{10.1145/3746027.3758206} provides global Mean Opinion Scores for rendered image quality, and MUGSQA~\cite{chen2026mugsqa} evaluates 3D scenes from diverse perspectives, including resolution and viewing distance. However, these methods remain limited to global quality perception and lack the potential to provide localized supervision for training. Furthermore, although Vision-Language Models (VLMs) demonstrate powerful semantic understanding, they lack spatial perception priors specific to 3D rendering artifacts, leading to weak spatial grounding and hallucinated evaluations. A unified, human-aligned, and fine-grained evaluation benchmark for 3DGS artifacts is urgently required.
% However, existing 3DGS quality assessment remains unable to answer these two questions.

% while failing to indicate the location and type of localized artifacts 要不要每个方法后面都写一句？

Conventional full-reference metrics, such as PSNR, SSIM~\cite{1284395}, and LPIPS~\cite{Zhang_2018_CVPR}, aggregate spatially varying errors into a single score, merely reflecting the global rendering quality. Recent 3DGS quality assessment datasets, including NVS-QA~\cite{zhang2025evaluatinghumanperceptionnovel}, GS-QA~\cite{11219925}, 3DGS-QA~\cite{wan2026perceptual}, 3DGS-IEval-15K~\cite{10.1145/3746027.3758206}, and MUGSQA~\cite{chen2026mugsqa}, broaden the evaluated scenes and perceptual dimensions, yet still rely primarily on image- or sequence-level quality annotations. Consequently, they cannot expose localized flaws hidden within otherwise high-quality renderings, nor do they provide the region masks and labels of artifacts for targeted refinement. General-purpose Vision-Language Models (VLMs) offer richer semantic descriptions, but their lack of 3DGS-specific artifact priors makes dense spatial grounding and fine-grained artifact discrimination unreliable. Existing approaches therefore remain limited to global quality prediction, leaving fine-grained diagnosis largely unexplored.

% To bridge this gap, we introduce \textbf{GaussFlaw-80K}, the first large-scale benchmark specifically designed for the fine-grained assessment of 3DGS rendering artifacts. As shown in Table~\ref{tab:dataset_comparison}, through a rigorous model-assisted, human-in-the-loop annotation pipeline, GaussFlaw-80K provides precise pixel-level region masks and degradation topology annotations for 80,000 distorted regions across 14,000 rendered images. We systematically establish a hierarchical taxonomy comprising 8 fine-grained topological categories (e.g., Transparent Haze, Ghosting, High-frequency Blur) derived from three high-level degradation dimensions. This benchmark fundamentally shifts the evaluation paradigm from implicit global averaging to explicit, region-level perceptual diagnosis, while simultaneously supporting further targeted feedback.

To address this gap, we introduce \textbf{GaussFlaw-80K}, the first large-scale dataset for fine-grained perceptual diagnosis in 3DGS with joint spatial and categorical annotations. Its pixel-level masks identify \textbf{where} perceptually significant artifacts occur, while its fine-grained labels characterize \textbf{what} type of flaw each region exhibits. Together, these annotations extend prior assessment from global score prediction to explicit flaw localization and classification (Table~\ref{tab:dataset_comparison}). GaussFlaw-80K contains 80,049 annotated artifact regions from 14,289 rendered images and organizes common 3DGS artifacts into a hierarchical taxonomy of eight categories spanning geometric, blending, and photometric degradations. The annotations are collected through a model-assisted human-in-the-loop pipeline to ensure an optimal trade-off between pixel-level precision and scalable annotation costs. By moving beyond image-level quality scores, GaussFlaw-80K establishes fine-grained artifact diagnosis as a measurable and reproducible task for 3DGS.

Building on the pixel-level supervision provided by GaussFlaw-80K, we propose \textbf{GeoCritic}, a geometry-guided multi-task perception model that jointly localizes 3DGS rendering artifacts and classifies their fine-grained flaw types. GeoCritic integrates RGB appearance with depth- derived geometric cues to improve spatial grounding and distinguish visually ambiguous artifacts. Beyond evaluation, we demonstrate the downstream benefit of fine-grained diagnosis for 3DGS refinement. Specifically, we translate GeoCritic's class-aware regional diagnosis into three complementary operations:  Gaussian densification for underrepresented content, attribute correction for unreliable primitives, and outlier pruning for detached Gaussians. Extensive experiments show that GeoCritic achieves state-of-the-art performance in both artifact localization and fine-grained classification (yielding an AUC-J of 0.9507 and an Acc@1 of 80.08\%). Furthermore, our targeted refinement significantly improves affected Gaussian representations across diverse datasets and backbones while minimizing unnecessary changes to well-reconstructed regions, demonstrating the practical value of region-level diagnosis for 3DGS refinement.

% These operations selectively improve the affected Gaussian representation while minimizing unnecessary changes to well-reconstructed regions, demonstrating the practical value of region-level diagnosis for 3DGS refinement.

Our contributions are summarized as follows:

\begin{itemize}
\item We introduce \textbf{GaussFlaw-80K}, the first dataset for fine-grained diagnosis of 3DGS rendering artifacts, providing \textbf{80,049} pixel-level artifact masks with category labels and enabling explicit region-level diagnosis beyond conventional global quality assessment.
% \item We introduce \textbf{GaussFlaw-80K}, the first fine-grained for 3DGS artifacts, providing over pixel-level topological annotations  to shift the evaluation paradigm from global averaging to explicit region-level diagnosis.
% We introduce \textbf{GaussFlaw-80K}, the first large-scale, fine-grained evaluation benchmark for 3DGS rendering artifacts. Featuring a rigorously established hierarchical taxonomy, it provides 80,000 precise pixel-level masks with topological annotations, fundamentally shifting the evaluation paradigm from implicit global averaging to explicit, region-level diagnosis.

\item We propose \textbf{GeoCritic}, a geometry-guided diagnostic expert that synergizes visual semantics with physical depth priors, enabling fine-grained spatial grounding and explicit topological reasoning of rendering artifacts.

\item  We demonstrate the downstream benefit of fine-grained diagnosis by translating class-aware regional predictions into three targeted Gaussian operations, which consistently improve local perceptual quality across multiple 3DGS backbones.

\item Extensive experiments validate the superiority of our proposed GeoCritic in accurately localizing and classifying 3DGS artifacts. Additionally, our empirical results demonstrate that the targeted refinement effectively rectifies local flaws, achieving noticeable perceptual improvements over existing baselines.

% \item Extensive experiments validate the effectiveness of our proposed framework. Our evaluation not only provides statistical insights into existing 3DGS paradigms but also demonstrates that our feedback-guided optimization significantly surpasses existing baselines in perceptual quality.

% \item Built upon the rich spatial and topological priors of GaussFlaw-80K, we present a systematic diagnosis of current 3DGS paradigms, revealing their capability boundaries and characteristic failure modes in rendering.

% \item We demonstrate the practical utility of our benchmark and diagnostic model through a set of targeted optimization strategies. Extensive experiments validate that leveraging GeoCritic's explicit region-level guidance precisely rectifies local artifacts without compromising global scene consistency, outperforming existing baselines.
\end{itemize}

\section{GaussFlaw-80K Construction}

\begin{figure*}[t]
  \centering
  \includegraphics[width=1.0\linewidth]{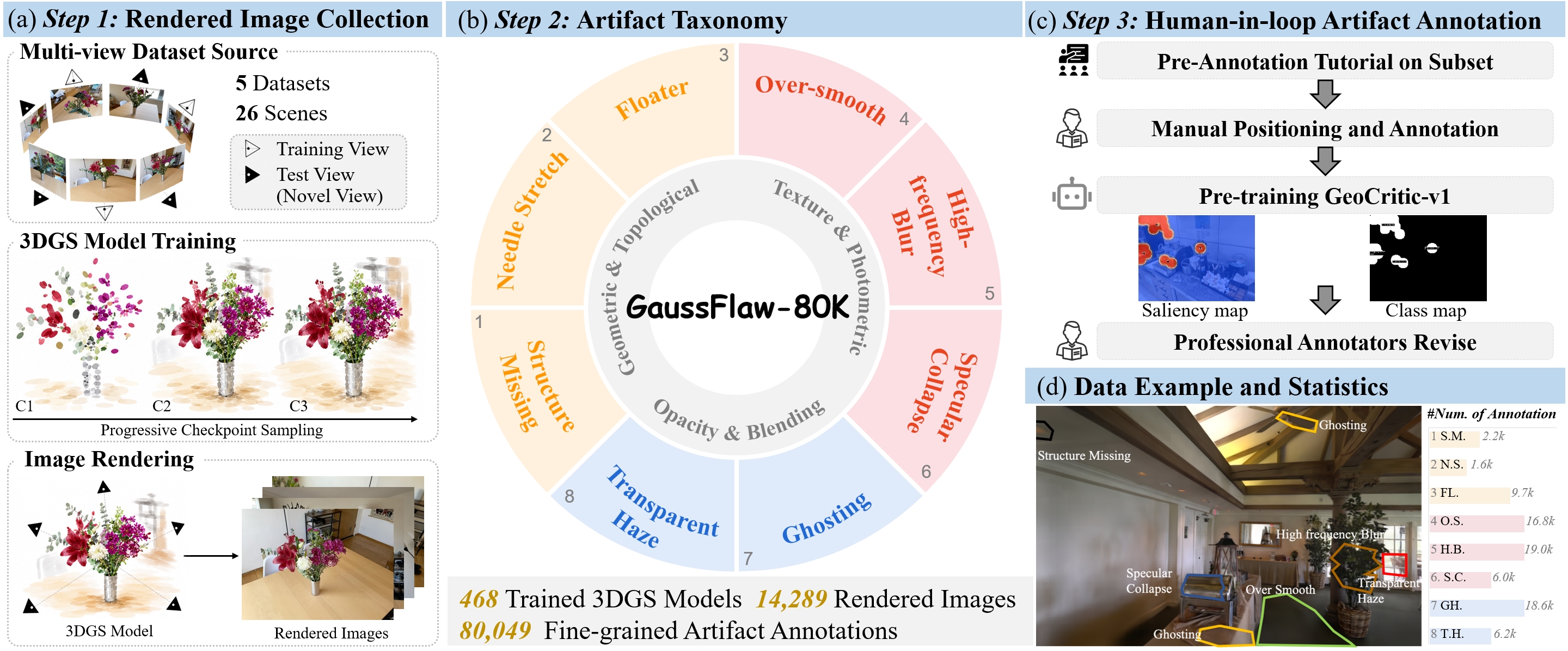}
  % \caption{Overview of GaussFlaw-80K.}
  \caption{\textbf{Data Construction Pipeline of GaussFlaw-80K.} (a) rendered image collection across datasets, 3DGS methods, checkpoints, and novel views; (b) the hierarchical taxonomy of eight artifact categories; (c) the human-in-the-loop annotation pipeline; and (d) annotated examples and dataset statistics.}
  \label{fig:1}
  % \vspace{-10pt}
\end{figure*}

We construct GaussFlaw-80K, a large-scale dataset specifically designed for fine-grained diagnosis of 3DGS rendering artifacts. It provides precise pixel-level region masks and fine-grained flaw type annotations for over 80K flaw regions across 8 artifact categories.

\subsection{Artifact Taxonomy}

As shown in Table~\ref{tab:dataset_comparison}, existing 3DGS evaluation datasets are primarily restricted to global quality scores (e.g., MOS) at a limited scale. To overcome this, we advance the annotation format from traditional holistic scoring to pixel-level diagnosis. Supported by an unprecedented volume of 80K annotations, this fine-grained paradigm enables precise spatial location and categorization of specific rendering flaws.

Central to this pixel-level assessment is a hierarchical taxonomy, as illustrated in Figure~\ref{fig:1}, which organizes common 3DGS artifacts into eight specific categories across three high-level dimensions. Specifically, \textit{Geometric and Topological Distortions} encompass \textbf{Floater}, characterized by detached or foreground-like blobs; \textbf{Needle Stretch}, manifested as thin and elongated spike-like structures; and \textbf{Structure Missing}, referring to absent or incomplete scene structures. \textit{Opacity and Blending Artifacts} include \textbf{Transparent Haze}, which appears as veil-like opacity accumulation over otherwise valid content, and \textbf{Ghosting}, characterized by duplicated, trailing, or semi-transparent structures. Lastly, \textit{Texture and Photometric Flaws} comprise \textbf{Over-smooth}, characterized by excessively smoothed structures and appearance; \textbf{High-frequency Blur}, referring to the loss of fine textures and edge details; and \textbf{Specular Collapse}, manifested as spurious, unstable, or distorted view-dependent highlights. At a high level, these artifacts reflect insufficient representation coverage, unreliable Gaussian attributes, and inadequate geometric support. This taxonomy provides a structured and interpretable foundation for fine-grained perceptual diagnosis.

\subsection{Data Collection and Annotation}

% \textbf{Rendering Sources and Diversity.} To ensure a comprehensive and generalizable data distribution, we curate rendering outputs from three representative 3DGS methods: vanilla 3DGS~\cite{kerbl20233d}, FastGS~\cite{Ren_2026_CVPR}, and HAC++~\cite{11106187}. Specifically, we train a total of 78 models across 26 sequences sourced from 5 typical datasets. To capture the diverse representations of artifacts across different training phases, we extract rendered images at six distinct checkpoints for each model, starting from 5,000 steps with an interval of 5,000 steps. Furthermore, for each sequence, we comprehensively sample multiple novel views covering a 360-degree range of the scene. This rigorous sampling strategy guarantees that GaussFlaw captures both iteration-dependent geometric anomalies and highly view-dependent degradations.

\textbf{Rendering Sources and Diversity.} To ensure a comprehensive and generalizable data distribution, we curate rendering outputs from three representative 3DGS methods: vanilla 3DGS~\cite{kerbl20233d}, FastGS~\cite{Ren_2026_CVPR}, and HAC++~\cite{11106187}. Specifically, we train models using three methods across 26 scenes sourced from five widely used datasets (Mip-NeRF 360~\cite{barron2022mip}, DeepBlending~\cite{10.1145/3272127.3275084}, LERF~\cite{lerf2023}, TandT~\cite{kerbl20233d}, and Tanks \& Temples~\cite{knapitsch2017tanks}). Recognizing that rendering artifacts dynamically evolve and manifest distinct visual characteristics at different training stages, we extract rendered images at six distinct checkpoints for each model, starting from 5K steps with a 5K-step interval. This temporal sampling yields a total of 468 intermediate checkpoint models. Furthermore, we comprehensively sample multiple novel views for each scene.

\textbf{Human-in-the-loop annotation pipeline.} We design a four-stage model-assisted human-in-the-loop annotation pipeline.
(1) Annotators first align with the established taxonomy through a pre-annotation phase. Subsequently, a small initial subset of rendered images is precisely annotated manually, providing pixel-level boundary masks and corresponding fine-grained topological labels for each distorted region.
(2) We pre-train a preliminary artifact perception model GeoCritic-v1 (detailed in Sec. 3) on this initial subset, and apply it to the remaining unannotated renderings to automatically generate candidate masks and topological pseudo-labels.
(3) Annotators manually adjust inaccurate mask boundaries, correct misclassified labels, filter out false-positive predictions, and add artifact regions missed by the model. 
(4) Final annotations are reconciled through majority voting and cross-validation by experts to eliminate subjective biases.
Consequently, each sample in GaussFlaw comprises a 3DGS rendering, precise artifact region masks, and fine-grained artifact labels. This paradigm significantly reduces annotation cost compared with fully manual annotation.

\textbf{Quality Control.} To ensure annotation quality, all GeoCritic-v1 proposals were manually reviewed and corrected. Starting from 17,444 manually annotated seed regions, GeoCritic-v1 generated 62,510 candidate regions, of which annotators retained 52,359. Among the retained masks, annotators refined the boundaries of 92.9\%, changed the areas of 39.99\% by more than 10\%, and corrected the category labels of 15.3\%. Annotators also added 10,246 missed regions. Expert cross-validation achieved over 95\% annotation agreement, demonstrating the high consistency and reliability of the final category labels and pixel-level masks. Finally, we obtained \textbf{80,049} pixel-level artifact regions with categorie labels across \textbf{14,289} rendered images, averaging \textbf{5.6} regions per image. As shown in Figure~\ref{fig:1}(d), High-frequency Blur and Ghosting are the most prevalent categories (23.7\% and 23.2\%, respectively), with further statistics provided in the supplementary material.

\begin{figure*}[t]
  \centering
  \includegraphics[width=1.0\linewidth]{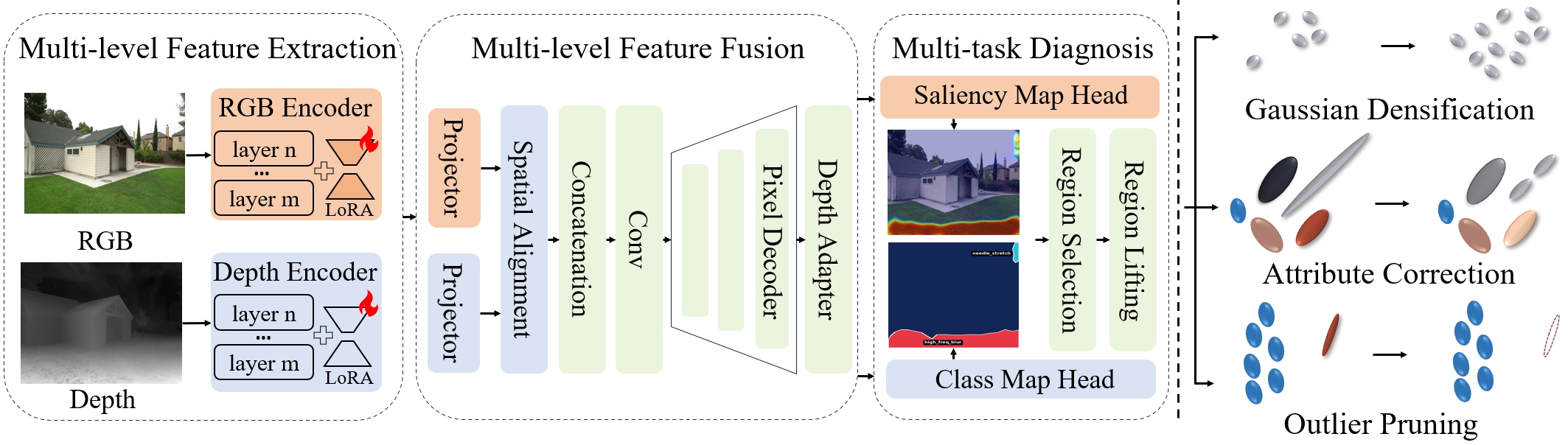}
  \caption{\textbf{Overview of the proposed fine-grained perceptual diagnosis pipeline.} Left: The GeoCritic model, leveraging a dual-stream architecture for RGB-depth feature extraction and multi-level fusion, followed by a multi-task head for fine-grained diagnosis. Right: The downstream targeted rectification operations guided by the selected artifact regions.}
  \label{fig:overview}
  \vspace{-10pt}
\end{figure*}

\section{Fine-Grained Perceptual Diagnosis}

\label{haha}

In this section, we first present \textbf{GeoCritic}, a multi-task perception model that simultaneously predicts dense artifact saliency and fine-grained flaw categories. We then introduce a downstream \textbf{Targeted Rectification} application, which projects GeoCritic's 2D predictions back to 3D Gaussians for precise geometric control (e.g., targeted expansion, attribute repair, and outlier removal) without compromising valid scene structures.

\subsection{GeoCritic: Geometry-Guided Artifact Perception}

% 3DGS renderings frequently exhibit complex distortions such as floaters, ghosts, and over-smoothing. These artifacts are often lack of distinct appearance boundaries, making conventional pure-RGB visual detection unreliable. To accurately diagnose these degradations, we design a geometry-guided perception model, which estimates both a distortion-saliency map $S \in [0, 1]^{H \times W}$ and a topological class $C$. As illustrated in left of Fig.~\ref{fig:overview}, a dual-stream architecture independently extracts hierarchical features from the RGB and Depth inputs. These multi-modal representations are subsequently integrated via a Multi-level Fusion module to capture the inherent physical correlations between visual appearance and spatial geometry. A Pixel Decoder then aggregates these fused multi-scale contextual cues to jointly predict the distortion saliency and classify the fine-grained degradation topology.

% 3D Gaussian Splatting (3DGS) renderings frequently exhibit spatially ambiguous and context-dependent degradations, such as floaters, ghosting, and needle-like stretching. These artifacts typically lack explicit object contours and only manifest as perceptual artifacts when local appearance is jointly interpreted with the underlying scene geometry. To explicitly diagnose these degradations, we design a geometry-guided dense perception model, which is systematically constructed from two core perspectives: geometry-guided representation learning and unified multi-task diagnostic prediction.
3DGS rendering artifacts typically exhibit ambiguous boundaries and complex categories. To explicitly diagnose these artifacts, we design a geometry-guided dense perception model, which is systematically constructed from two core perspectives: geometry-guided representation learning and unified multi-task diagnostic prediction.

\textbf{Geometry-Guided Representation Learning.} As illustrated in the left panel of Figure~\ref{fig:overview}, the model leverages a dual-stream architecture to inject physical geometric priors. The appearance branch extracts rich semantic features $\{R_i\}_{i=1}^{4}$ via an RGB encoder, while the geometry branch employs a depth encoder to capture multi-level depth cues $\{G_i\}_{i=1}^{4}$. To seamlessly integrate these two modalities, we introduce a lightweight multi-level fusion module. At each hierarchical level, the RGB and depth features are mapped into a shared latent space via projection layers. Following spatial alignment and channel-wise concatenation, the combined multi-modal representations are integrated through a convolutional fusion:
\begin{equation}
    F_i = \mathcal{F}_i \Big( Concat\big[ \mathcal{P}^r_i(R_i), \text{A}(\mathcal{P}^d_i(G_i)) \big] \Big)
\end{equation}
where $\mathcal{P}^r_i$ and $\mathcal{P}^d_i$ denote the projection operations for RGB and depth, respectively, $\text{A}(\cdot)$ represents the spatial alignment, and $\mathcal{F}_i$ stands for the convolutional fusion layer. The fused features $\{F_i\}_{i=1}^{4}$ are aggregated by an FPN-style pixel decoder. We further introduce a CNN-based Depth Adapter that injects depth-derived boundary cues into decoded features, enforcing geometry-aware localization and suppressing false positives in physically consistent regions.

\textbf{Unified Multi-Task Diagnostic Prediction.} Instead of cascading separate localization and classification networks, we attach a unified multi-task head to the shared dense features, enabling the joint diagnosis of artifact regions and their corresponding topological flaw types. From the decoded representation, the unified head predicts an artifact saliency map $S \in [0,1]^{H \times W}$ and a fine-grained class map $C \in [0,1]^{K \times H \times W}$. This design tightly couples spatial localization with category-aware evidence through shared dense features and aligned supervision. The network is optimized via a dual supervision objective:
\begin{equation}
     \mathcal{L} =\lambda_c \mathcal{L}_{class} + \lambda_s \mathcal{L}_{saliency}
\end{equation}

For the artifact saliency branch, we supervise the saliency prediction with a fidelity term and a distribution-alignment term:
\begin{equation}
    \mathcal{L}_{saliency}
    =
    \mathcal{L}_{fid}(S, S^\ast)
    +
    \mathcal{L}_{dist}(\hat{S}, \hat{S}^\ast),
\end{equation}
where $S^\ast$ denotes the ground-truth artifact saliency map. Given the inherently ambiguous visual boundaries of 3DGS artifacts, we apply a Gaussian kernel to the pixel-level annotations to generate this continuous ground-truth saliency map. $\hat{S}$ and $\hat{S}^\ast$ denote the corresponding normalized saliency distributions. The fidelity term encourages local response consistency with the annotated artifact regions, while the distribution-alignment term encourages the predicted saliency distribution to match the human-annotated artifact saliency density. This formulation supports both pixel-level localization fidelity and distribution-level saliency consistency. For the class heatmap branch, we apply a per-class sigmoid focal loss:
  \begin{equation}
      \mathcal{L}_{class}
      =
      \mathcal{L}_{focal}(C, C^\ast),
  \end{equation}
where $C^\ast$ is the ground-truth class map. During inference, the
predicted saliency map $S$ is thresholded to extract connected components as localized artifact proposals. The flaw type of each proposal is then obtained by aggregating the corresponding spatial responses in the class map $C$.

\subsection{Downstream Application: Targeted Rectification}
%After obtaining pixel-level artifact maps and topological predictions from GeoCritic, GaussFlaw transfers the class-aware evidence from the image plane to the Gaussians that repeatedly contribute to the affected regions across views. This association reveals whether a local defect stems from insufficient representation capacity, unreliable primitive attributes, or geometrically unsupported primitives. GaussFlaw then expands missing capacity, repairs unreliable primitives, or removes confirmed outliers, while leaving well-reconstructed regions unchanged.
%After obtaining class-aware artifact maps from GeoCritic, GaussFlaw associates the image-space diagnoses with the underlying 3D representation. We project the Gaussians into multiple probe views and aggregate their class-specific responses together with visibility, depth consistency, and surface-support evidence. This process identifies whether a diagnosed region reflects insufficient representation capacity, unreliable Gaussian attributes, or geometrically unsupported primitives. GaussFlaw accordingly performs three complementary forms of localized refinement: Gaussian densification, attribute correction, and outlier pruning.

After obtaining class-aware artifact maps from GeoCritic, GaussFlaw translates the image-space diagnoses into 3D refinement targets through two steps: region selection and region lifting. Region selection extracts connected artifact regions from the predicted maps and assigns each region its corresponding degradation type. Region lifting then associates the selected regions with the underlying Gaussians by projecting the primitives into multiple probe views and aggregating their class-specific responses across views under visibility, depth-consistency, and surface-support constraints. Based on the lifted evidence, GaussFlaw applies three complementary forms of localized refinement: Gaussian densification, attribute correction and outlier pruning.

%\textbf{Expanding insufficient coverage.} Some artifacts indicate that the current representation lacks the capacity to explain local scene content. For example, high-frequency blur, over-smoothing, and missing structures often arise when fine surfaces or texture-bearing geometry are underrepresented. GaussFlaw addresses this limitation by introducing additional Gaussian seeds in diagnostically identified regions before optimization. The new primitives are initialized from multi-view consistent surface evidence, so that the added capacity follows the existing scene geometry instead of being distributed uniformly throughout the scene. This expansion mechanism allows the model to allocate more representation to regions where details are genuinely absent, improving local completeness without unnecessarily increasing the complexity of already reliable areas.
\begin{figure*}[t]
  \centering
  \includegraphics[width=1.0\linewidth]{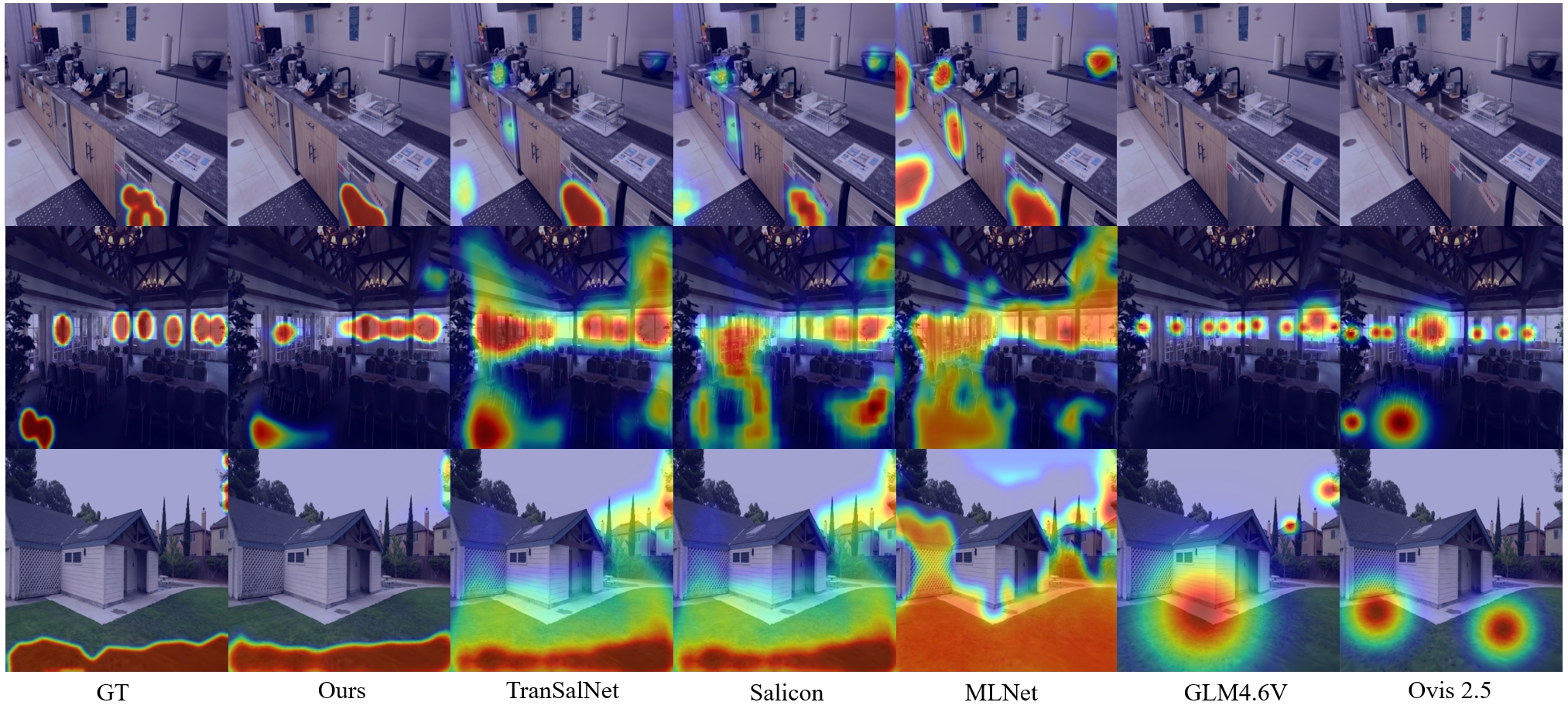}
  \caption{Qualitative comparison of artifact location results. Our method yields sharper and more precise localization than deep saliency detectors and VLM-based methods. More qualitative results are included in the supplementary material.}
  \label{fig:qualitive_saliency}
  \vspace{-10pt}
\end{figure*}
\textbf{Gaussian Densification.} High-frequency blur, over-smoothing, and missing structures often indicate insufficient local representation capacity. For these regions, modifying existing Gaussians alone may be ineffective because the required geometry is underrepresented or absent. GaussFlaw establishes multi-view correspondences within the diagnosed regions and triangulates depth-consistent observations into additional surface seeds. These seeds augment the original initialization before re-optimization, allocating new representation capacity to geometrically supported regions containing missing structures or fine details.

%\textbf{Repairing unreliable primitives.} Other artifacts are not caused by missing primitives, but by Gaussians whose attributes no longer provide a faithful explanation of the scene. Transparent haze, ghosting, needle-like stretching, and appearance collapse are representative examples: the underlying geometry may remain useful, whereas its opacity, scale, or view-dependent appearance is inaccurate. GaussFlaw therefore repairs these primitives in a category-aware manner, adjusting the attributes most closely related to the diagnosed degradation. Local refinement can subsequently be increased around reliable repair targets to recover details that may have been suppressed by the original parameterization. In this way, the model corrects the source of the artifact while preserving valid geometry that would otherwise be lost under an overly aggressive removal strategy.

%\textbf{Attribute Correction.} Artifacts such as transparent haze, ghosting, needle-like stretching, and specular collapse may arise from Gaussians with plausible geometry but unreliable attributes. GaussFlaw therefore performs category-aware correction of their opacity, anisotropic scale, or view-dependent appearance. It further performs repair-conditioned Gaussian splitting around reliable correction targets when additional local capacity is required. This operation preserves useful geometry while correcting the attributes responsible for the diagnosed artifact.

\textbf{Attribute Correction.} GaussFlaw corrects unreliable attributes of geometrically plausible Gaussians in a category-specific manner: opacity is reduced for transparent haze, ghosting, and non-detached floaters; the largest scale dimension is contracted for needle stretching and anisotropic ghosting; and higher-order SH coefficients are attenuated for specular collapse. Gaussian positions and unrelated attributes remain unchanged, while the corresponding optimizer states are reset to preserve the corrections during subsequent training.

%\textbf{Removing unsupported outliers.} Pruning is used only when the diagnosis indicates that a primitive is unlikely to belong to the physical scene. In particular, a floater prediction alone is not sufficient for deletion, since visually ambiguous regions can still correspond to valid surfaces. GaussFlaw removes a Gaussian only when its artifact evidence is accompanied by inconsistent depth behavior and weak multi-view surface support, which together indicate a detached or unsupported outlier. This conservative rule separates truly spurious primitives from uncertain but recoverable regions, which remain available for repair. Consequently, GaussFlaw avoids the common trade-off of global pruning methods, where removing distracting floaters can also damage thin structures or view-dependent content.

\textbf{Outlier Pruning.}  GaussFlaw projects the diagnosed floater and needle-stretch regions back to the 3D representation to identify the Gaussians responsible for these artifacts. It then aggregates their evidence across views and compares their projected depths with the corresponding rendered surfaces. Gaussians that consistently appear as foreground outliers without sufficient surface support are regarded as detached from the scene. Only these verified outliers are removed from the Gaussian representation under a conservative pruning budget, while all other candidates are excluded from pruning.

\section{Experiments}

%\subsection{Experimental Setup}

%\textbf{Datasets and Implementation Details.} We evaluate PACER on both 3DGS rectification and artifact perception. For rectification, experiments are conducted on representative 3D reconstruction datasets, including Mip-NeRF 360[cite], Tanks \& Temples[cite], and xxx[cite]. To demonstrate the broad applicability of PACER, we implement it across three architectures: vanilla 3DGS, FastGS, and HAC++. For perception, we train and evaluate our model on GaussFlaw-80K. Both encoders are adapted with LoRA using rank 8 and alpha = 32. The model is trained for 40 epochs using AdamW with a learning rate of 3e-5, weight decay of 1e-4, and batch size 8.

%\textbf{Evaluation Metrics.} For rectification, we report standard full-image fidelity metrics, including PSNR[cite], SSIM[cite], LPIPS[cite], and DISTS[cite]. Since localized artifacts may be obscured by global averaging, we additionally compute crop-level metrics: PSNR$_c$, SSIM$_c$, and LPIPS$_c$, over artifact regions. FID[cite] and CMMD[cite] are further adopted to measure distributional perceptual consistency. For artifact perception, we follow saliency evaluation protocols and report AUC-Judd[cite], NSS[cite], CC[cite], SIM[cite], and KLD[cite] for artifacts localization. We also report Acc@1, Acc@3, and Acc@5 for fine-grained topological classification. Detailed training settings are provided in the supplementary material.

\subsection{Implementation Details}

\textbf{GeoCritic.} We implement GeoCritic using pretrained SigLIP 2~\cite{tschannen2025siglip2multilingualvisionlanguage} and Depth Anything 3 encoders~\cite{depthanything3} for appearance and geometry representation, respectively. Both encoders are adapted with LoRA using a rank of 16. For the depth adapter, we employ a lightweight CNN that simultaneously processes the dense features output by the pixel decoder and the normalized depth gradients. GeoCritic is optimized for 60 epochs using AdamW with a learning rate of $2\times10^{-5}$, a weight decay of $1\times10^{-4}$, and a batch size of 8. The loss weights are set to $\lambda_s= 1.5$ and $\lambda_c = 1$. Training is conducted on 2 NVIDIA A100 GPUs.

\textbf{Targeted Rectification.}  For diagnosis-guided rectification, we first train a short preliminary reconstruction to generate the additional surface seeds used for Gaussian densification. We add at most 12,000 seeds per scene and initialize them with a scale multiplier of 0.35 and an opacity of 0.15. The final reconstruction is optimized for 30K iterations, with GeoCritic invoked at iterations 8K, 15K, and 23K. During attribute correction, opacity values are multiplied by 0.65 for transparent haze, ghosting, and non-detached floaters; the largest scale dimension is multiplied by 0.75 for needle stretching and by 0.90 for anisotropic ghosting; and higher-order SH coefficients are multiplied by 0.70 for specular collapse. GeoCritic is used only to guide these explicit operations, and no auxiliary perceptual loss is introduced during reconstruction. All remaining optimization settings follow the official configuration of each 3DGS backbone. 

\begin{table}[t]
\centering
% 核心修改 1：加入 \small 让表格字体整体比正文小一号
\small 

\renewcommand\arraystretch{1.2}
% 核心修改 2：因为字变小了，列间距可以稍微放宽一点到 3.5pt，让版面更舒展
\setlength{\tabcolsep}{2.5pt} 
\begin{tabular}{l ccccc}
\toprule[1pt]
\textbf{Model} & \textbf{AUC-J$\uparrow$} & \textbf{NSS$\uparrow$} & \textbf{CC$\uparrow$} & \textbf{SIM$\uparrow$} & \textbf{KLD$\downarrow$} \\ \midrule[0.8pt]

% ==================== Traditional Methods ====================
SR            & 0.4322 & -0.1132 & -0.0839 & 0.2575 & 2.4141 \\
PFT           & 0.4370 & -0.1070 & -0.0794 & 0.2651 & 2.1804 \\
SWD           & 0.4493 & -0.2018 & -0.0753 & 0.2998 & 2.0255 \\
SUN           & 0.4249 & -0.1704 & -0.1036 & 0.2827 & 2.1543 \\
Murray        & 0.4726 &  0.0234 & -0.0287 & 0.3043 & 2.1249 \\ \cdashline{1-6}

% ==================== Deep Saliency Methods ====================
Salicon       & 0.9419 & 1.9878  & 0.6745  & 0.5305 & 1.0941 \\
MLNet         & 0.9286 & 1.6901  & 0.5935  & 0.4580 & 1.2734 \\
Sal-CFS-GAN   & 0.8415 & 1.6022  & 0.5703  & 0.5030 & 3.0521 \\
SAM-VGG       & 0.9355 & 2.0124  & 0.6607  & 0.5298 & 1.1257 \\
SAM-ResNet    & 0.9251 & 1.9043  & 0.6274  & 0.4958 & 1.2108 \\
TranSalNet    & 0.9436 & 2.0629  & 0.6744  & 0.5424 & \textbf{1.0486} \\ \cdashline{1-6}

% ==================== MLLM Baselines ====================
Qwen3-VL-8B              & 0.5138 & 0.0210 & 0.0094 & 0.0332 & 8.3581 \\
Qwen3-VL-8B$^\dagger$    & 0.7164 & 0.8402 & 0.3240 & 0.3381 & 2.3794 \\
GLM-4.6V-9B              & 0.4976 & 0.0012 & 0.0007 & 0.1537 & 3.0515 \\
GLM-4.6V-9B$^\dagger$    & 0.7193 & 0.9241 & 0.3170 & 0.3165 & 3.0042 \\
Ovis 2.5-9B               & 0.4435 & -0.0738 & -0.0307 & 0.0706 & 7.0901 \\
Ovis 2.5-9B$^\dagger$     & 0.6114 & 0.3488 & 0.1306 & 0.1862 & 4.5628 \\
Gemma 4-E4B           & 0.5283   & 0.0508 & 0.0189 & 0.0525 & 7.5783 \\
Gemma 4-E4B$^\dagger$ & 0.7059 & 0.7062 & 0.2502 & 0.2689 & 3.9959 \\ \midrule[0.8pt]

% ==================== Proposed Methods ====================
Ours (RGB-only)      & \underline{0.9493} & \underline{2.1390} & \underline{0.6987} & \underline{0.6095} & 1.1085 \\
\textbf{Ours (RGB+depth)} & \textbf{0.9507} & \textbf{2.1680} & \textbf{0.7062} & \textbf{0.6201} & \underline{1.0898} \\ \bottomrule[1pt]
\end{tabular}
\caption{Quantitative evaluation of the GeoCritic for artifact location. Unmarked VLMs are zero-shot, and $\dagger$ denotes LoRA fine-tuning. Best and second-best results are shown in bold and underlined, respectively.}
\label{tab:saliency_eval}
\vspace{-10pt}
\end{table}

\begin{table}[t]
\centering

\small % 保持较小的字号
\renewcommand\arraystretch{1.15} % 稍微增加行高，让表格在高度上更饱满一点
\setlength{\tabcolsep}{3.5pt} % 控制列间距来调整表格宽度
\begin{tabular}{l cccc}
\toprule[1pt]
\textbf{Model} & \textbf{Acc@1$\uparrow$} & \textbf{Acc@3$\uparrow$} & \textbf{M-Recall$\uparrow$} & \textbf{M-F1$\uparrow$} \\ \midrule[0.8pt]

Qwen3-VL-8B              & 27.56\% & 76.32\% & 0.1775 & 0.1479 \\
Qwen3-VL-8B$^\dagger$    & 75.37\% & 83.46\% & 0.6596 & 0.6748 \\ 
GLM-4.6V-9B              & 14.39\% & 50.00\% & 0.1106 & 0.0776 \\
GLM-4.6V-9B$^\dagger$    & 75.11\% & 82.87\% & 0.6230 & 0.6455 \\
Ovis 2.5-9B              & 16.23\% & 54.60\% & 0.1054 & 0.0872 \\
Ovis 2.5-9B$^\dagger$    & 67.12\% & 77.54\% & 0.5371 & 0.5407 \\
Gemma 4-E4B              & 10.86\% & 39.55\% & 0.0927 & 0.0638   \\
Gemma 4-E4B$^\dagger$    & 71.07\% & 80.45\% & 0.5859 & 0.6021   \\ \midrule[0.8pt]

Ours (RGB-only) & \underline{78.97\%} & \underline{95.74\%} & \underline{0.6877} & \underline{0.7048} \\
\textbf{Ours (RGB+depth)} & \textbf{80.08\%} & \textbf{97.60\%} & \textbf{0.7235} & \textbf{0.7285} \\ \bottomrule[1pt]
\end{tabular}
\caption{Quantitative evaluation of GeoCritic for fine-grained topological artifact classification. Unmarked VLMs are zero-shot, and $\dagger$ denotes LoRA fine-tuning. Best and second-best results are shown in bold and underlined, respectively.}
\label{tab:classification_eval}
\vspace{-10pt}
\end{table}

\begin{table*}[t]                                                                                                                                                                                                                                 
  \centering                                                                                                                                                                                                                                        
  \small                                                                                                                                                                                                                                            
  \renewcommand\arraystretch{1.30}                                                                                                                                                                                                                  
  \setlength{\tabcolsep}{2.5pt}                                                                                                                                                                                                                     
                                                                                                                                                                                                                                                    
  \begin{tabular}{l cccc cc cccc cc}                                                                                                                                                                                                                
  \toprule[1pt]                                                                                                                                                                                                                                     
  \multirow{3}{*}{\textbf{Model}}                                                                                                                                                                                                                   
  & \multicolumn{6}{c}{\textbf{Mip-NeRF 360}}                                                                                                                                                                                                       
  & \multicolumn{6}{c}{\textbf{TandT}} \\                                                                                                                                                                                                           
  \cmidrule(lr){2-7} \cmidrule(lr){8-13}                                                                                                                                                                                                            
  & \multicolumn{4}{c}{\textbf{Global Quality}}                                                                                                                                                                                                     
  & \multicolumn{2}{c}{\textbf{Region Quality}}                                                                                                                                                                                                     
  & \multicolumn{4}{c}{\textbf{Global Quality}}                                                                                                                                                                                                     
  & \multicolumn{2}{c}{\textbf{Region Quality}} \\                                                                                                                                                                                                  
  \cmidrule(lr){2-5} \cmidrule(lr){6-7}                                                                                                                                                                                                             
  \cmidrule(lr){8-11} \cmidrule(lr){12-13}                                                                                                                                                                                                          
  & \textbf{PSNR$\uparrow$}                                                                                                                                                                                                                         
  & \textbf{SSIM$\uparrow$}                                                                                                                                                                                                                         
  & \textbf{LPIPS$\downarrow$}                                                                                                                                                                                                                      
  & \textbf{DISTS$\downarrow$}                                                                                                                                                                                                                      
  & \textbf{LPIPS$_c$$\downarrow$}                                                                                                                                                                                                                  
  & \textbf{DISTS$_c$$\downarrow$}                                                                                                                                                                                                                  
  & \textbf{PSNR$\uparrow$}                                                                                                                                                                                                                         
  & \textbf{SSIM$\uparrow$}                                                                                                                                                                                                                         
  & \textbf{LPIPS$\downarrow$}                                                                                                                                                                                                                      
  & \textbf{DISTS$\downarrow$}                                                                                                                                                                                                                      
  & \textbf{LPIPS$_c$$\downarrow$}                                                                                                                                                                                                                  
  & \textbf{DISTS$_c$$\downarrow$} \\                                                                                                                                                                                                               
  \midrule[0.8pt]                                                                                                                                                                                                                                   
                                                                                                                                                                                                                                                    
  Vanilla 3DGS                                                                                                                                                                                                                                      
  & 27.749 & 0.823 & 0.176 & 0.069 & 0.231 & 0.148                                                                                                                                                                                                  
  & 23.826 & 0.853 & 0.169 & 0.063 & 0.222 & 0.137 \\                                                                                                                                                                                               
  \hfill \textbf{+ Ours}                                                                                                                                                                                                                            
  & \textbf{27.895} & \textbf{0.828} & \textbf{0.166} & \textbf{0.065} & \textbf{0.206} & \textbf{0.134}                                                                                                                                            
  & \textbf{23.959} & \textbf{0.858} & \textbf{0.159} & \textbf{0.060} & \textbf{0.201} & \textbf{0.128} \\                                                                                                                                         
  \cdashline{1-13}

  Scaffold-GS
  & 28.430 & 0.833 & 0.170 & 0.070 & 0.223 & 0.144
  & 23.991 & 0.850 & 0.174 & 0.068 & 0.207 & 0.132 \\
  \hfill \textbf{+ Ours}
  & \textbf{28.448} & \textbf{0.838} & \textbf{0.158} & \textbf{0.063} & \textbf{0.195} & \textbf{0.126}
  & \textbf{24.379} & \textbf{0.857} & \textbf{0.160} & \textbf{0.063} & \textbf{0.180} & \textbf{0.117} \\
  \cdashline{1-13}

  Octree-GS
  & 28.165 & 0.833 & 0.148 & 0.068 & 0.216 & 0.144
  & 24.178 & 0.856 & 0.163 & 0.064 & 0.198 & 0.128 \\
  \hfill \textbf{+ Ours}
  & \textbf{28.267} & \textbf{0.836} & \textbf{0.140} & \textbf{0.065} & \textbf{0.194} & \textbf{0.131}
  & \textbf{24.414} & \textbf{0.858} & \textbf{0.155} & \textbf{0.062} & \textbf{0.177} & \textbf{0.118} \\
  \bottomrule[1pt]
  \end{tabular}

  \caption{Quantitative evaluation of GaussFlaw-guided rectification across different 3DGS backbones. Metrics are reported on both complete rendered images (Global) and diagnosed-region crops (Region).}
  \label{tab:rectification_comparison}
  \vspace{-5pt}
  \end{table*}

\subsection{GeoCritic Analysis}

\textbf{Experimental Setup.} For artifact localization, following~\cite{8315047}, we adopt CC, SIM, KLD,
AUC-Judd, and NSS to evaluate pixel-wise localization and distributional consistency. For topological artifact classification, we employ Top-1 and Top-3 accuracy (Acc@1, Acc@3), alongside Macro-Recall (M-Recall) and Macro-F1 (M-F1). While Acc@1 strictly evaluates primary label accuracy, we introduce Acc@3 as a relaxed metric to reflect predictive robustness given the inherent visual ambiguity of 3DGS artifacts. Furthermore, M-Recall and M-F1 are computed as the macro-average across all 8 flaw categories, which are incorporated to further measure the stability of the model's recognition capability across different classes.

\textbf{Comparisons.} Table~\ref{tab:saliency_eval} and Table~\ref{tab:classification_eval} evaluate the proposed GeoCritic in terms of artifact localization and topological artifact classification. Conventional saliency detectors (SR~\cite{4270292}, PFT~\cite{4587715}, SWD~\cite{5995676}, SUN~\cite{SUN}, and Murray~\cite{5995506}) perform poorly because they primarily respond to low-level contrast rather than rendering-specific distortions. Deep saliency networks (Salicon~\cite{7410395}, MLNet~\cite{mlnet2016}, Sal-CFS-GAN~\cite{Che_2020}, SAM~\cite{cornia2018predicting}, and TranSalNet~\cite{LOU2022455}) achieve better localization results, but their lack of geometry priors limits their ability to recognize 3DGS-specific artifacts even after task-specific training. Moreover, these single-task saliency methods inherently lack the capability to classify topological artifact categories. General-purpose VLMs (Qwen3-VL-8B~\cite{bai2025qwen3vltechnicalreport}, GLM-4.6V-9B~\cite{vteam2025glm45vglm41vthinkingversatilemultimodal}, Ovis 2.5-9B~\cite{lu2025ovis25technicalreport}, and Gemma 4-E4B~\cite{gemmateam2026gemma4technicalreport}) exhibit limited artifact grounding in the zero-shot setting. Even with fine tuning, these VLMs remain limited in both localization and classification, suggesting that general multimodal priors are insufficient for fine-grained 3DGS artifact diagnosis.

By contrast, GeoCritic delivers the strongest overall performance across the two tasks (\textbf{AUC-J = 0.9507, CC = 0.7062, SIM = 0.6201, Acc@1 = 80.08\%}), demonstrating the effectiveness of our Unified Multi-Task Diagnostic Prediction, which jointly models \textit{where} rendering artifacts occur and \textit{what} flaw types they correspond to. To further assess Geometry-Guided Representation Learning, we compare GeoCritic with an RGB-only variant that removes depth-derived geometric cues. The full model consistently improves both localization and classification performance over this variant, increasing AUC-J from 0.9493 to 0.9507 and Acc@1 from 78.97\% to 80.08\%. These gains show that depth-derived geometry provides complementary evidence to RGB appearance, leading to more reliable artifact diagnosis.

Figure~\ref{fig:qualitive_saliency} presents qualitative visualizations of artifact localization by GeoCritic. Compared with deep saliency networks (producing coarse responses with many false positives) and VLMs (tending to miss subtle artifacts and provide imprecise spatial grounding), GeoCritic delivers sharper spatial focus and stronger agreement with ground-truth annotations. Such precise localization provides a reliable foundation for fine-grained artifact diagnosis.

\subsection{Targeted Rectification Analysis}
\textbf{Experimental Setup.} We evaluate diagnosis-guided rectification on Mip-NeRF 360~\cite{barron2022mip} and TandT~\cite{kerbl20233d} using three representative 3DGS backbones: 3DGS~\cite{kerbl20233d},  Scaffold-GS~\cite{scaffoldgs} and Octree-GS~\cite{10993308}. For each backbone, we compare its official baseline with the corresponding GaussFlaw-guided variant while retaining the original reconstruction objective, optimizer, and backbone-specific training configuration. We report PSNR, SSIM~\cite{1284395}, LPIPS~\cite{Zhang_2018_CVPR}, and DISTS~\cite{ding2020iqa} on complete renderings. To evaluate localized artifacts that may be obscured by global averaging, we additionally report LPIPS$_c$ and DISTS$_c$ on diagnosed-region crops. The crop boundaries are determined once from the baseline predictions and shared by both variants; the masks only define these boundaries, and all pixels within each crop are retained for metric computation.

\begin{table}[t]
\centering
\small
\renewcommand{\arraystretch}{1.2}
\setlength{\tabcolsep}{3.5pt}

\begin{tabular}{l cccc}
\toprule[1pt]
\textbf{Metric}
& \textbf{Baseline}
& \textbf{Densification}
& \textbf{Correction}
& \textbf{Pruning} \\
\midrule[0.8pt]

\textbf{LPIPS$_c$$\downarrow$}
& 0.231
& 0.212
& 0.225
& 0.229 \\

\textbf{DISTS$_c$$\downarrow$}
& 0.148
& 0.138
& 0.144
& 0.146 \\

\bottomrule[1pt]
\end{tabular}
\caption{Quantitative analysis of individual diagnosis-guided operations on Mip-NeRF 360 with Vanilla 3DGS. Metrics are computed over diagnosed-region crops.}
\label{tab:individual_operations}
\vspace{-10pt}
\end{table}

\textbf{Comparisons across 3DGS Backbones.} Table~\ref{tab:rectification_comparison} demonstrates the consistent effectiveness and backbone-agnostic applicability of GaussFlaw. Under full-image evaluation, GaussFlaw consistently improves PSNR and SSIM while reducing LPIPS and DISTS across all six dataset-backbone combinations, demonstrating stable improvements in both global reconstruction fidelity and perceptual quality. More importantly, its benefits are particularly evident within the diagnosed artifact regions. On Mip-NeRF 360, GaussFlaw reduces the average LPIPS$_c$ and DISTS$_c$ by \textbf{0.025} and \textbf{0.015}, respectively. Consistent local improvements are obtained on TandT, with average reductions of \textbf{0.023} in LPIPS$_c$ and \textbf{0.011} in DISTS$_c$. Notably, both crop-level perceptual metrics improve for every backbone on both datasets, demonstrating that GaussFlaw consistently enhances perceptually degraded regions across different Gaussian representations.   

\begin{figure}[t]
  \centering
  \includegraphics[width=1.0\linewidth]{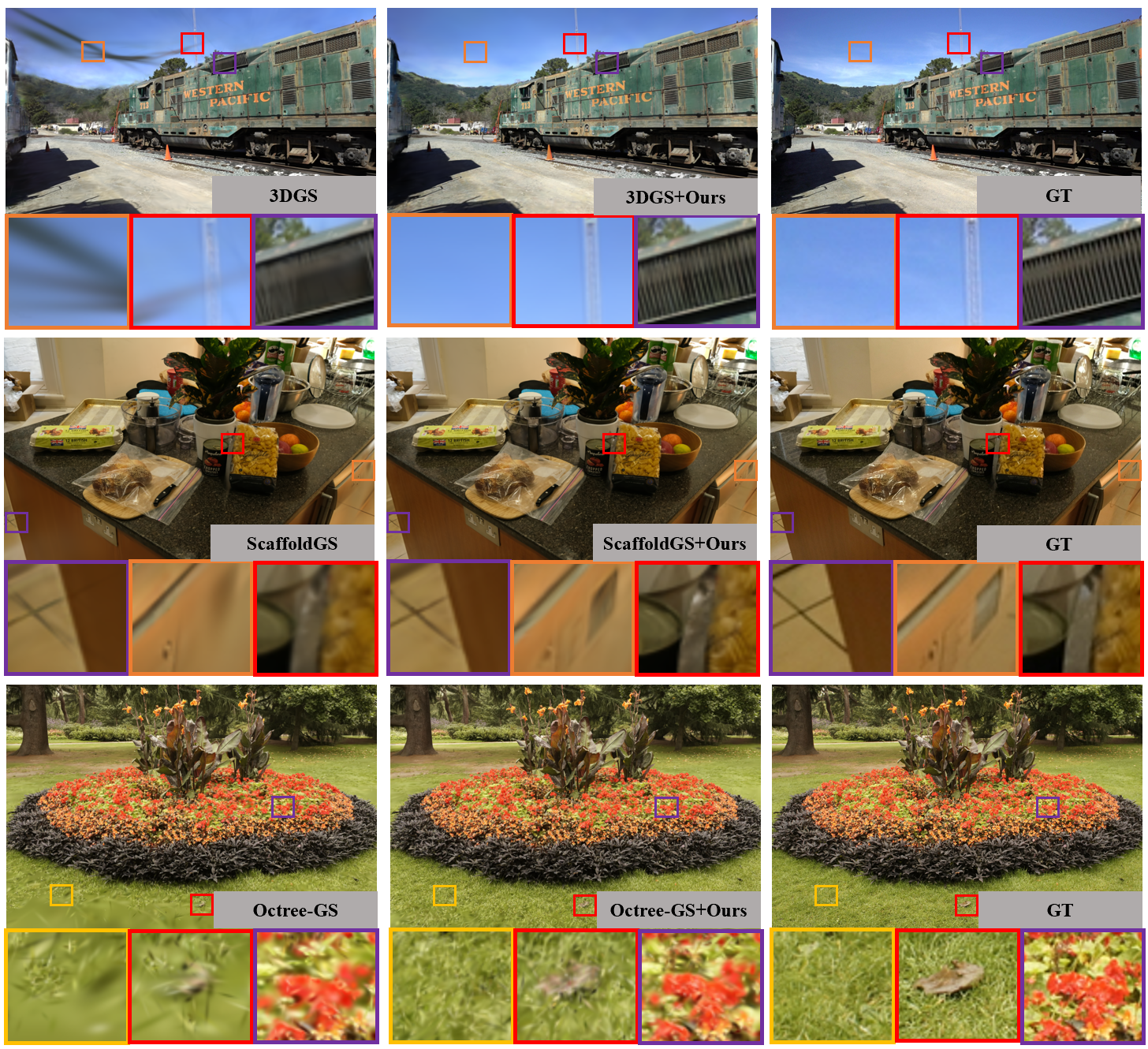}
  \caption{Qualitative comparison of GaussFlaw-guided rectification across   
  different 3DGS backbones.}
  \label{fig:qualitative}
  \vspace{-10pt}
\end{figure}
Qualitative comparisons (Figure~\ref{fig:qualitative}) further corroborate these improvements, particularly in challenging and perceptually sensitive regions such as skies, vegetation, floors, finely textured surfaces, and reflective objects. GaussFlaw restores coherent fine-scale structures, suppresses over-smoothing, haze, and geometric stretching, and removes detached floaters without disturbing well-reconstructed content. These results further demonstrate that GeoCritic provides reliable and actionable guidance for targeted 3DGS rectification.

\textbf{Analysis of Individual Operations.} To disentangle the contribution of each diagnosis-guided operation, we conduct controlled ablations by enabling Gaussian densification, attribute correction and outlier pruning individually under the same backbone and training configuration. As shown in Table~\ref{tab:individual_operations}, each operation consistently improves all crop-level metrics over the baseline, confirming its independent contribution to targeted rectification. Among them, Gaussian densification yields the largest overall improvement, reducing LPIPS$_c$ and DISTS$_c$ by \textbf{0.019} and \textbf{0.010}, respectively, highlighting its effectiveness in recovering missing structures and high-frequency details. Attribute correction and outlier pruning also improve local fidelity and perceptual quality, validating the effectiveness of correcting unreliable Gaussian attributes and removing geometrically unsupported primitives.

\section{Conclusion}
We introduced GaussFlaw-80K, a dataset for fine-grained perceptual diagnosis of rendering artifacts in 3D Gaussian Splatting. Building on this dataset, we proposed GeoCritic, which combines appearance and geometric cues to jointly localize artifacts and classify their flaw types. We further demonstrated the practical value of these diagnoses through three targeted rectification operations that directly refine the affected Gaussians. Experiments confirm GeoCritic's reliable diagnostic performance and the consistent global and local improvements achieved by diagnosis-guided refinement. These results highlight fine-grained perceptual diagnosis as a promising direction for interpretable and actionable 3DGS optimization.
% \input{sec/7_Acknowledgement}


\bigskip
%\noindent Thank you for reading these instructions carefully. We look forward to receiving your electronic files!

\bibliography{aaai2027}


\end{document}